\documentclass[10pt,twocolumn,letterpaper]{article}

\usepackage{iccv}
\usepackage{times}
\usepackage{epsfig}
\usepackage{graphicx}
\usepackage{amsmath}
\usepackage{amssymb}
\usepackage{afterpage}

\usepackage[pagebackref=true,breaklinks=true,letterpaper=true,colorlinks,bookmarks=false]{hyperref}

\iccvfinalcopy 

\def\iccvPaperID{****} 

\ificcvfinal\fi

\begin{document}

\title{On-device Real-time Hand Gesture Recognition}

\author{
George Sung\quad Kanstantsin Sokal\quad Esha Uboweja\quad Valentin Bazarevsky\quad Jonathan Baccash\\
Eduard Gabriel Bazavan\quad Chuo-Ling Chang\quad Matthias Grundmann\\
Google Research\\
1600 Amphitheatre Pkwy, Mountain View, CA 94043, USA\\
{\tt\small \{gsung, kanstantsin, eshauboweja, bazarevsky, jbaccash, egbazavan, chuoling, grundman\}@google.com}\\
}

\maketitle
\ificcvfinal\thispagestyle{empty}\fi

\begin{abstract}
   We present an on-device real-time hand gesture recognition (HGR) system, which detects a set of predefined static gestures from a single RGB camera. The system consists of two parts: a hand skeleton tracker and a gesture classifier. We use MediaPipe Hands~\cite{zhang20, mediapipe_hands} as the basis of the hand skeleton tracker, improve the keypoint accuracy, and add the estimation of 3D keypoints in a world metric space. We create two different gesture classifiers, one based on heuristics and the other using neural networks (NN).
\end{abstract}

\section{Introduction}

Hand gesture recognition (HGR) is a natural and intuitive method for human-computer interaction (HCI), and has been an active research area~\cite{rautaray2015vision,oudah2020hand}. A wide variety of input devices and techniques have been investigated, and skeleton-based HGR is a popular choice due to its robustness to background and light variations~\cite{devineau2018deep,de2016skeleton}. Many skeleton-based HGR systems rely on depth sensors, such as RGBD cameras, which are not nearly as common as RGB cameras on mobile devices. Our HGR, on the other hand, requires only a single RGB camera. It does so by first predicting 3D skeleton keypoints from a camera image, then running a gesture classifier on the keypoints.

We design two gesture classifiers with different use cases in mind. The heuristics-based classifier is easier to create and extend, without the need for training data, and more intuitive to develop and troubleshoot. The NN-based classifier is more accurate and precise, especially for borderline cases. It’s also more forgiving of errors in skeleton keypoints.

Our HGR runs real-time at 30fps on mainstream mobile devices. 

\section{Architecture}

\begin{figure}
\begin{center}
   \includegraphics[width=1\linewidth]{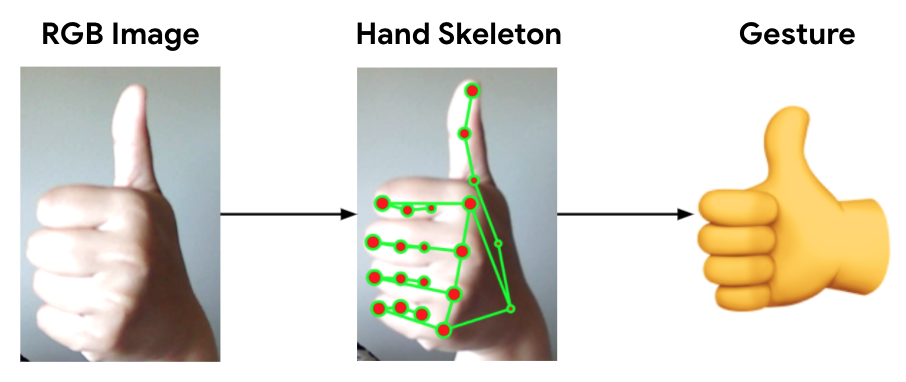}
\end{center}
   \caption{Our hand gesture recognition system}
\label{fig:architecture}
\end{figure}

Our HGR consists of two parts: a hand skeleton tracker improved from MediaPipe Hands and a gesture classifier, as shown on Figure \ref{fig:architecture}. The two-step approach has a few advantages:
\begin{itemize}
    \item Reduced engineering effort by leveraging the hand tracker which is already real-time, robust, and fair~\cite{hands_model_card}.
    \item Simpler gesture classifier design which processes skeleton keypoints instead of raw pixels.
    \item Optimized complexity by running the gesture classifier only when hands are tracked.
\end{itemize}

\subsection{Hand Skeleton Tracker}\label{section:hand_skeleton_tracker}

Both our gesture classifiers operate on a keypoint level (not with RGB data), therefore accurate hand keypoint estimation is a key prerequisite for gesture classification. As a basis for our gesture recognition pipeline, we improve MediaPipe Hands~\cite{mediapipe_hands} by training with a new sophisticated hand poses data (like American Sign Language). Analysis of various gestures indicates that robust estimation of hand rotation angle and normalization distance is key to an accurate hand tracker. The original hand rotation and scale estimation is based on the 2D vector from the wrist to the middle finger knuckle. For various cases (like frontal view) such normalization becomes very unstable as the normalization distance approaches zero, which results in a significant degradation in tracking quality. To overcome this problem, we introduce a new algorithm similar to the approach taken by BlazePose ~\cite{blazepose2020}. 

\begin{figure}
\begin{center}
   \includegraphics[width=1\linewidth]{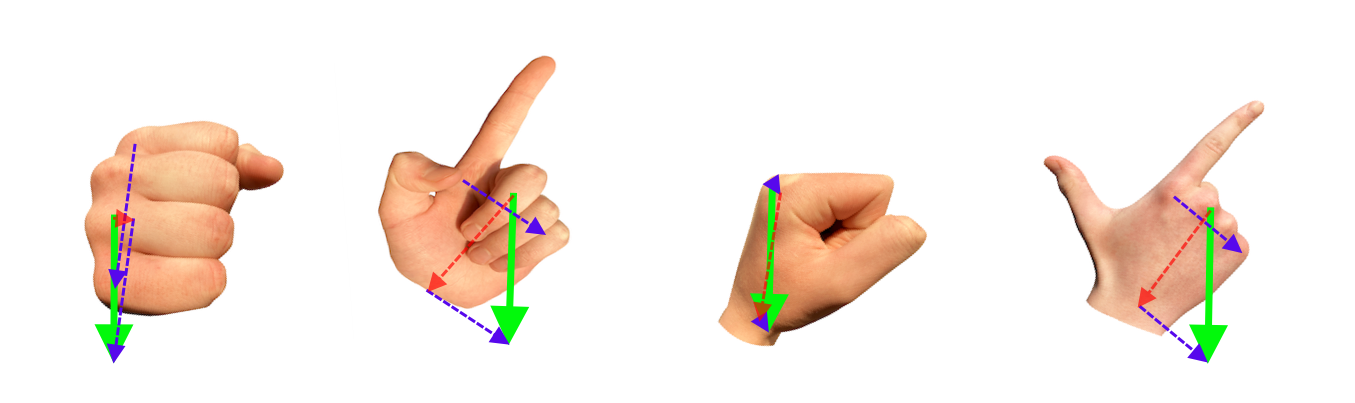}
\end{center}
   \caption{Hands rotation angle derived from the sum of two vectors: index to pinky base knuckle (in green) and middle base knuckle to wrist (in red).}
\label{fig:hand_alignment}
\end{figure}

For our case, we define two virtual keypoints to describe hand center, scale and rotation angle: a center keypoint and an alignment keypoint. The center keypoint is estimated as the average of the index, middle and pinky knuckles.
The alignment keypoint location is estimated so it forms the rotation/scale vector with the center keypoint. The rotation angle is estimated from a sum of two vectors: from the middle base knuckle to the wrist, and from the index to the pinky base knuckle. As the component vectors tend to be orthogonal in the majority of cases; the resulting sum vector changes smoothly for any hand pose, and never degrades to zero, as shown on Figure \ref{fig:hand_alignment}. This increases overall hand tracking quality for frontal hand cases. The scalar value of the alignment vector is estimated as the distance from the center keypoint to the farthest knuckle of the same hand. The new rotation and scale normalization results in a significant quality boost for the whole hand pose estimation pipeline: 71.3 mAP vs 66.5 mAP (for the original MediaPipe Hands ~\cite{mediapipe_hands} pipeline) on our validation dataset with complex ASL hand poses.

\begin{figure}
\begin{center}
   \includegraphics[width=1\linewidth]{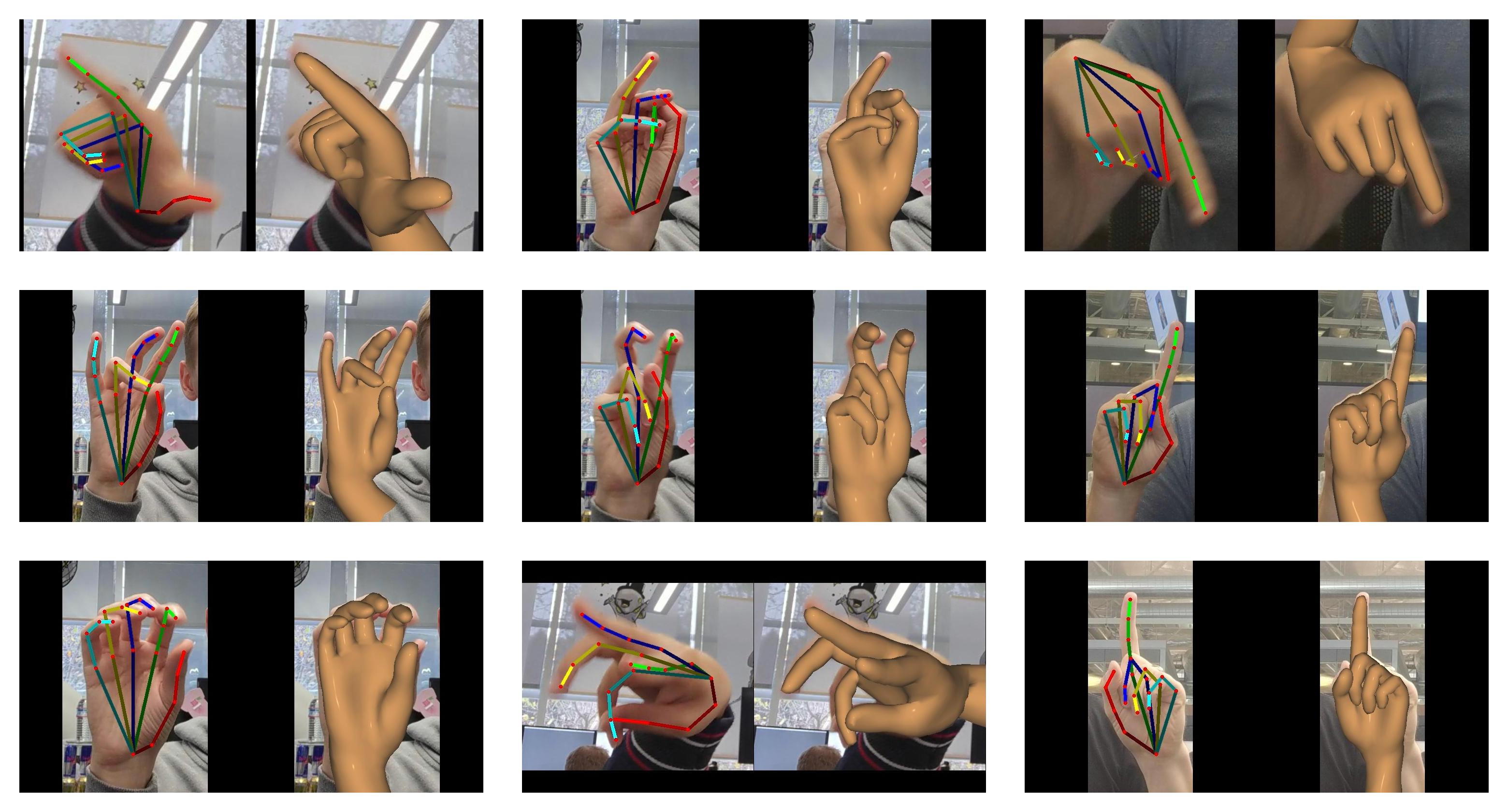}
\end{center}
   \caption{Sample images with overlaid 2D annotations and the corresponding GHUM \cite{xu2020ghum} hand models fitted and rendered on top.}
\label{fig:ghum}
\end{figure}

Accurate hand pose estimation in the 3D space is a vital component for both angle based and few-shot learning gesture classification. It minimizes the ambiguity among projections of the same hand pose from different observer positions in space, and allows the gesture classification to be invariant to rotation. Therefore, in addition to predicting the hand pose in the screen ``pixel'' space, we also estimate the pose in a world metric space relatively to the hand wrist. To obtain a 3D hand pose ground truth in a metric space, we fit our 2D hand annotation with a statistical and highly realistic GHUM ~\cite{xu2020ghum,zanfir2020weakly} hand model as shown in Figure \ref{fig:ghum}. Due to the nature of perspective projection, objects of different sizes may have the same projection on the image plane: two objects with the same shape (bigger and smaller) will have the same projection if placed respectively further and closer to the camera. Therefore we have to make the following assumptions when we perform the hand fitting from 2D keypoints: human hands have minor variations in size, those variations are always reflected in hand shapes, and thus covered by the statistical GHUM hand model. For images with unknown camera intrinsic parameters, we assume the focal length is the maximum of the image width and height, and the optical center is the center of the image. Since both the model training and inference operate on a cropped image, we normalize the absolute world coordinates such that the origin is at the middle finger knuckle, while the scale remains the same. We also normalize 3D coordinates in a roll plane using virtual keypoints to be consistent with input image transformations.

Overall, our hand tracking model achieves a mean average prediction error of 1.5cm, and provides 21 hand keypoints in a metric 3D space to the gesture classfier.

\subsection{Heuristics Gesture Classifier} \label{section:heuristic_gesture} 

Based on the hand skeleton tracker, we build a single-shot, heuristics-based classifier for a small set of static gestures. We start with the simple gesture classification approach described in~\cite{zhang20}, which first derives a set of angles between various 2D hand keypoints, then applies thresholds to the derived angles to define a discrete state for each finger (e.g. bent or straight), and finally defines a static gesture as a logic expression based on the finger states. To get a more accurate set of underlying angles, we replace features based on the 2D hand keypoints with features based on the 3D world metric hand keypoints. In order to de-correlate the gesture classifier features and make manual threshold picking easier, we distinguish between \textit{extrinsic} and \textit{intrinsic} features with respect to the palm pose in the 3D world metric space during the preprocessing stage:

\begin{itemize}
\setlength{\parskip}{0pt}
\setlength{\itemsep}{0pt plus 1pt}
    \item The \textit{extrinsic} features are composed of palm pose components such as rotation, scale and translation. For classification purposes, we only use the rotation pose component of the palm represented by its three Euler angles.
    \item The \textit{intrinsic} features are angles between various hand keypoints. For classification purposes, we derive a single feature angle for each finger with a goal of thresholding them to define a discrete state (e.g. fully bent, fully straight or neither). In accordance with the underlying hand skeleton topology, each finger is represented by a base joint keypoint, two intermediate joint keypoints and a tip keypoint. To define an individual finger feature angle, we first introduce a 3D polygonal chain: starting from the wrist keypoint, to the finger's base joint keypoint, the first intermediate joint keypoint, the second intermediate joint keypoint, and finishing at the tip keypoint. Then, we define the feature angle as the maximum angle between the first chain segment and each of the remaining chain segments. Additionally, we derive a feature angle for each adjacent pair of fingers with the same goal of thresholding them to define a discrete state (e.g. fingers crossed, apart or neither).
\end{itemize}

\begin{figure}
\begin{center}
   \includegraphics[width=1\linewidth]{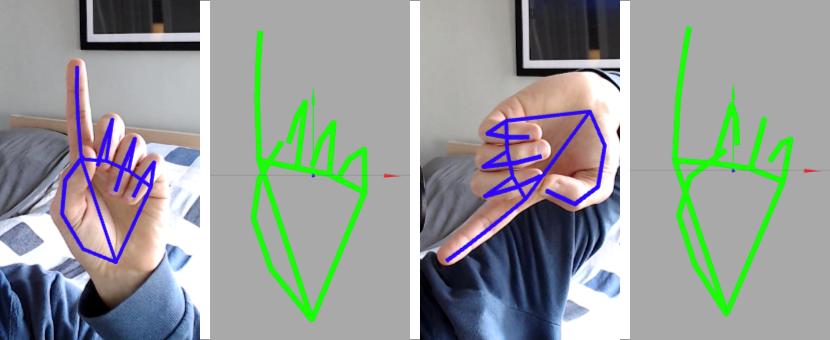}
\end{center}
   \caption{3D hand keypoints decoupled from the palm pose during the preprocessing stage. The blue hand skeleton is based on the 2D hand keypoints. The green hand skeleton is based on the preprocessed 3D hand keypoints.}
\label{fig:heuristic_extrinsic_vs_intrinsic}
\end{figure}

As shown on Figure \ref{fig:heuristic_extrinsic_vs_intrinsic}, removing the influence of the extrinsic features from the the intrinsic features produces a consistent 3D hand keypoint set for a fixed hand shape configuration,  regardless of its position on the input frames.  In turn, this significantly de-correlates features and makes it manageable to manually pick thresholds on a larger feature set in order to define a more complex static gesture.

\begin{figure}
\begin{center}
   \includegraphics[width=1\linewidth]{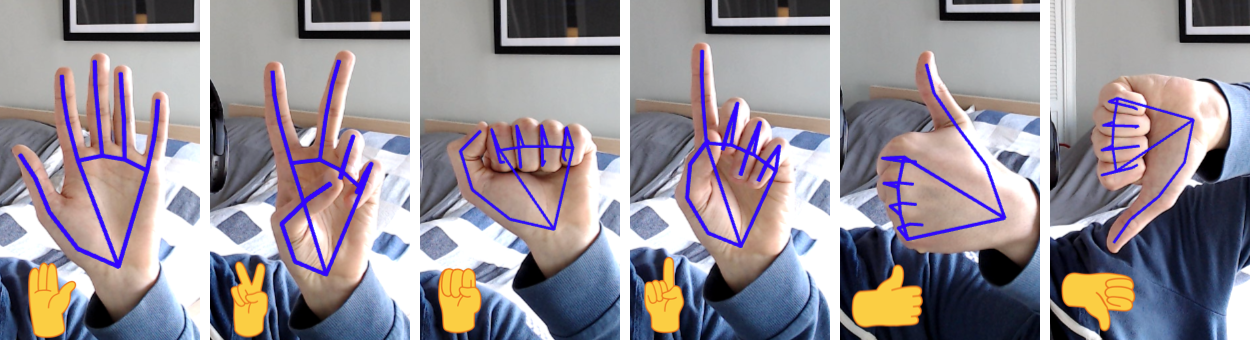}
\end{center}
   \caption{Visualization of gestures supported by the heuristic-based classifier. Left-to-right: \textit{OpenPalm}, \textit{Victory}, \textit{ClosedFist}, \textit{PointingUp}, \textit{ThumbUp}, \textit{ThumbDown}.}
\label{fig:heuristic_example}
\end{figure}

As the final stage of the heuristics-based classifier, we establish a system of 6 gesture definitions based on the features derived from the pre-processed 3D world metric hand keypoints. Figure \ref{fig:heuristic_example} showcases the supported gestures. This particular gesture set is chosen so that it covers some of the most recognizable and common static hand gestures.

To evaluate our approach, we collect and annotate an in-house gesture dataset. The dataset contains 1882 short video clips that cover various angles and lighting conditions for 21 static hand gestures (for the full list of gestures, please see Appendix A). The dataset contains all 6 gestures supported by the heuristics-based classifier (see Figure \ref{fig:heuristic_example}) and those are used as positive samples. The remaining 15 gestures are used as negative samples. The presence of a diverse negative sample collection allows us to evaluate how often the classifier recognizes an unknown gesture as a known one. The limitation of this dataset is that it’s collected from only 18 users with limited variation in background. The classifier achieves 0.86\% false positive rate and 44.4\% recall rate on the dataset.

\subsection{NN Gesture Classifier}\label{section:nn_gesture}

\begin{figure}
\begin{center}
  \includegraphics[width=1\linewidth]{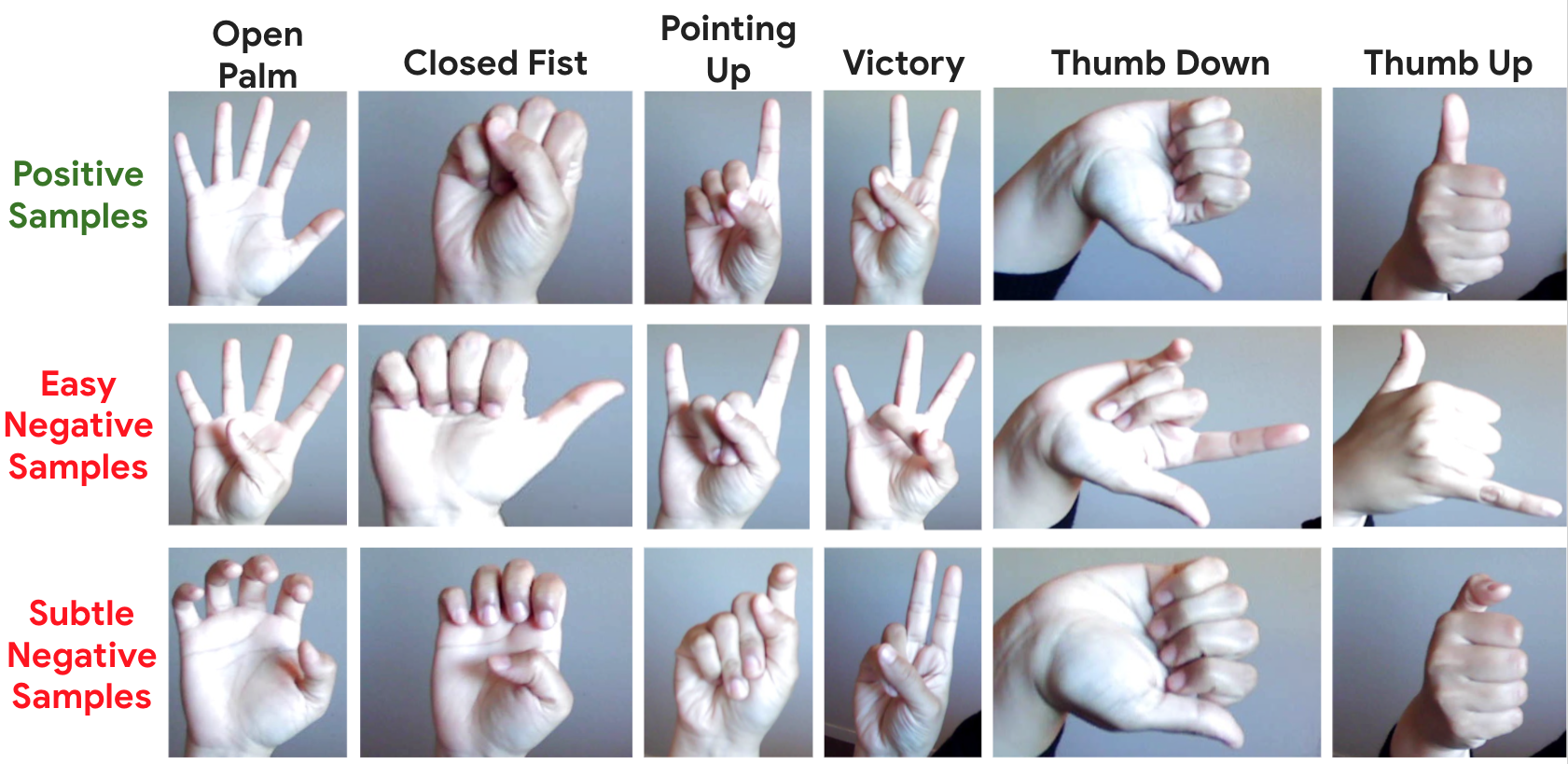}
\end{center}
  \caption{Some examples of true positive samples for gesture classes, easy samples for \textit{Negative} hand shapes and subtle variations of hand shapes that should not be confused with the gesture class.}
\label{fig:dataset_samples}
\end{figure}

The dataset used in Section~\ref{section:heuristic_gesture} is for evaluation only and too small for training NN models.
We collect and mine another in-house dataset containing $7307$ images from $6478$ users, representing a rich variety of hand shapes of both gestures and non-gestures in the wild. In addition to regular positive samples we collect easy and hard \textit{Negative} samples. Please see Figure~\ref{fig:dataset_samples} for some example images.

We train an NN classifier to distinguish among six static hand gestures, namely, \textit{OpenPalm}, \textit{ClosedFist}, \textit{PointingUp}, \textit{Victory}, \textit{ThumbUp}, \textit{ThumbDown} and a background \textit{Negative} class.  The NN model consists of $3$ fully connected layers of $50$ neurons each. The model inputs are the intrinsic and extrinsic features computed in Section~\ref{section:heuristic_gesture}. We use focal loss~\cite{lin2017focal} to deal with the class imbalance, where there are a lot more negative samples than positive samples in our dataset.

The NN classifier achieves an average recall rate of $87.9\%$ across the $6$ static gesture classes at a false positive rate of $1\%$.

\section{MediaPipe Implementation}\label{section:mediapipe}

The proposed HGR system is implemented using the open source MediaPipe framework~\cite{mediapipe_dev}. The system adds to MediaPipe Hands, primarily consisting of a hand-detection and a hand-keypoint component~\cite{zhang20, mediapipe_hands}, an additional gesture-classification component as discussed in Section~\ref{section:heuristic_gesture} and~\ref{section:nn_gesture}.

In many applications, such as remote control, the user gestures only once in a while but the HGR is always running in the background. In order to reduce average computation requirement and maintain real-time performance across a wide range of devices, the HGR system hand detection is configured to run only as needed, capped at a maximum frequency lower than the frequency of hand keypoint generation and gesture classification, by utilizing the flow-control and stream-synchronization support in MediaPipe~\cite{mediapipe} similar to~\cite{zhang20}. When there’s no hand in the camera view, the HGR runs hand detection at a lower frequency to save computation and power. As soon as a hand is detected, it’s tracked at a higher frequency for better accuracy and temporal resolution. Furthermore, GPU acceleration is heavily exploited end-to-end, covering tasks like ML model inference as well as image and tensor processing, via OpenGL/OpenCL/Metal on mobile devices and WebGL locally in Web browsers (similar to the Web ML effort enabling background blur/replace in Google Meet~\cite{segmentation_aiblog}).

\section{Applications}

Our HGR can be used as an HCI mechanism for various applications, such as a virtual touchscreen for desktop computers, sending visual commands for robots, a controller for virtual reality gaming systems, and a remote control for large screen displays~\cite{rautaray2015vision}.

{\small
\bibliographystyle{ieee_fullname}
\bibliography{egbib}

\begin{thebibliography}{10}\itemsep=-1pt

\bibitem{mediapipe_dev}
{MediaPipe}.
\newblock \url{https://mediapipe.dev/}, 2019.

\bibitem{mediapipe_hands}
{MediaPipe Hands}.
\newblock \url{https://solutions.mediapipe.dev/hands}, 2019.

\bibitem{segmentation_aiblog}
{Background Features in Google Meet, Powered by Web ML}.
\newblock
  \url{https://ai.googleblog.com/2020/10/background-features-in-google-meet.html},
  2020.

\bibitem{hands_model_card}
{MediaPipe Hands Model Card}.
\newblock \url{https://mediapipe.page.link/handmc}, 2020.

\bibitem{blazepose2020}
Valentin Bazarevsky, Ivan Grishchenko, Karthik Raveendran, Tyler Zhu, Fan
  Zhang, and Matthias Grundmann.
\newblock {BlazePose: On-device Real-time Body Pose Tracking}, 2020.

\bibitem{de2016skeleton}
Quentin De~Smedt, Hazem Wannous, and Jean-Philippe Vandeborre.
\newblock Skeleton-based dynamic hand gesture recognition.
\newblock In {\em Proceedings of the IEEE Conference on Computer Vision and
  Pattern Recognition Workshops}, pages 1--9, 2016.

\bibitem{devineau2018deep}
Guillaume Devineau, Fabien Moutarde, Wang Xi, and Jie Yang.
\newblock Deep learning for hand gesture recognition on skeletal data.
\newblock In {\em 2018 13th IEEE International Conference on Automatic Face \&
  Gesture Recognition (FG 2018)}, pages 106--113. IEEE, 2018.

\bibitem{lin2017focal}
Tsung-Yi Lin, Priya Goyal, Ross Girshick, Kaiming He, and Piotr Doll{\'a}r.
\newblock Focal loss for dense object detection.
\newblock In {\em Proceedings of the IEEE international conference on computer
  vision}, pages 2980--2988, 2017.

\bibitem{mediapipe}
Camillo Lugaresi, Jiuqiang Tang, Hadon Nash, Chris McClanahan, Esha Uboweja,
  Michael Hays, Fan Zhang, Chuo-Ling Chang, Ming~Guang Yong, Juhyun Lee,
  Wan-Teh Chang, Wei Hua, Manfred Georg, and Matthias Grundmann.
\newblock {MediaPipe: A Framework for Building Perception Pipelines}.
\newblock In {\em {CVPR Workshop on Computer Vision for AR/VR}}, 2019.

\bibitem{oudah2020hand}
Munir Oudah, Ali Al-Naji, and Javaan Chahl.
\newblock Hand gesture recognition based on computer vision: a review of
  techniques.
\newblock {\em journal of Imaging}, 6(8):73, 2020.

\bibitem{rautaray2015vision}
Siddharth~S Rautaray and Anupam Agrawal.
\newblock Vision based hand gesture recognition for human computer interaction:
  a survey.
\newblock {\em Artificial intelligence review}, 43(1):1--54, 2015.

\bibitem{xu2020ghum}
Hongyi Xu, Eduard~Gabriel Bazavan, Andrei Zanfir, William~T Freeman, Rahul
  Sukthankar, and Cristian Sminchisescu.
\newblock {GHUM \& GHUML: Generative 3D Human Shape and Articulated Pose
  Models}.
\newblock In {\em Proceedings of the IEEE/CVF Conference on Computer Vision and
  Pattern Recognition}, pages 6184--6193, 2020.

\bibitem{zanfir2020weakly}
Andrei Zanfir, Eduard~Gabriel Bazavan, Hongyi Xu, William~T. Freeman, Rahul
  Sukthankar, and Cristian Sminchisescu.
\newblock {Weakly Supervised 3D Human Pose and Shape Reconstruction with
  Normalizing Flows}.
\newblock In {\em Computer Vision -- ECCV 2020}, pages 465--481, 2020.

\bibitem{zhang20}
Fan Zhang, Valentin Bazarevsky, Andrey Vakunov, George Sung, Chuo-Ling Chang,
  Matthias Grundmann, and Andrei Tkachenka.
\newblock {MediaPipe Hands: On-device Real-time Hand Tracking}.
\newblock In {\em CVPR Workshop on Computer Vision for Augmented and Virtual
  Reality, Seattle, WA}, 2020.

\end{thebibliography}
}

\newpage
\section*{Appendix A. In-house static gesture dataset: gesture code names} \label{sect:appendixa}

\vspace{\topsep}
\begin{enumerate}
\setlength{\parskip}{0pt}
\setlength{\itemsep}{0pt plus 3pt}
  \item OpenPalm
  \item Victory
  \item ClosedFist
  \item PointingUp
  \item ThumbUp
  \item ThumbDown
  \item OK
  \item CallMe
  \item IndexMiddlePointingUp
  \item Three
  \item Four
  \item ILoveYou
  \item FingerHeart
  \item HandHeart
  \item IndexMiddlePointingUpWithClosedThumb
  \item IndexMiddlePointingUpWithOpenThumb
  \item IndexPointingToCamera
  \item Loser
  \item PinchedFingers
  \item VulcanSalute
  \item SignOfTheHorns
\end{enumerate}

\end{document}